# Parallel Replacement in Finite State Calculus

## André Kempe and Lauri Karttunen

Rank Xerox Research Centre – Grenoble Laboratory

6, chemin de Maupertuis – 38240 Meylan – France

{kempe,karttunen}@xerox.fr    http://www.rxrc.xerox.com/grenoble/mltt/home.html

## Abstract

This paper extends the calculus of regular expressions with new types of replacement expressions that enhance the expressiveness of the simple replace operator defined in Karttunen (1995). Parallel replacement allows multiple replacements to apply simultaneously to the same input without interfering with each other. We also allow a replacement to be constrained by any number of alternative contexts. With these enhancements, the general replacement expressions are more versatile than two-level rules for the description of complex morphological alternations.

## 1  Introduction

A replacement expression specifies that a given symbol or a sequence of symbols should be replaced by another one in a certain context or contexts. Phonological rewrite-rules (Kaplan and Kay, 1994), two-level rules (Koskenniemi 1983), syntactic disambiguation rules (Karlsson et al 1994, Koskenniemi, Tapanainen, and Voutilainen 1992), and part-of-speech assignment rules (Brill 1992, Roche and Schabes 1995) are examples of replacement in context of finite-state grammars.

Kaplan and Kay (1994) describe a general method representing a replacement procedure as finite-state transduction. Karttunen (1995) takes a somewhat simpler approach by introducing to the calculus of regular expression a replacement operator that is defined just in terms of the other regular expression operators. We follow here the latter approach.

In the regular expression calculus, the replacement operator, `->`, is similar to crossproduct, in that a replacement expression describes a relation between two simple regular languages. Consequently, regular expressions can be conveniently combined with other kinds of coperations, such as composition and union to form complex expressions.

A replacement relation consists of pairs of strings that are related to one another in the manner sketched below:

$$\begin{array}{lllll} x & u_i^j & y & u_i^k & z \quad \text{upper string} \quad [1] \\ x & l_i^j & y & l_i^k & z \quad \text{lower string} \end{array}$$

We use $u_i^j$ and $u_i^k$ to represent instances of $U_i$ (with $i \in [1, n]$) and $l_i^j$ and $l_i^k$ to represent instances of $L_i$. The upper string contains zero or more instances of $U_i$, possibly interspersed with other material (denoted here by x, y, and z). In the corresponding lower string the sections corresponding to $U_i$ are instances of $L_i$, and the intervening material remains the same (Karttunen, 1995).

The `->` operator makes the replacement obligatory, `(->)` makes it optional. For the sake of completeness, we also define the inverse operators, `<-`, and `(<-)`, and the bidirectional variants, `<->` and `(<->)`.

We have incorporated the new replacement expressions into our implementation of the finite-state calculus (Kempe and Karttunen, 1995). Thus, we can construct transducers directly from replacement expressions as part of the general calculus, without invoking any special rule compiler.

### 1.1  Simple regular expressions

The table below describes the types of regular expressions and special symbols that are used to define the replacement operators.

| | | |
|---|---|---|
| `(A)` | option, `[ A | 0 ]` | [2] |
| `A*` | Kleene star | |
| `A+` | Kleene plus | |
| `A/B` | ignore (A possibly interspersed with strings from B) | |
| `~A` | complement (negation) | |
| `$A` | contains (at least one) A | |
| `A  B` | concatenation | |
| `A | B` | union | |
| `A & B` | intersection | |
| `A - B` | relative complement (minus) | |
| `A .x. B` | crossproduct (Cartesian product) | |
| `A .o. B` | composition | |
| `0 or [ ]` | epsilon (the empty string) | |
| `[. .]` | affects empty string replacement (sec. 2.2) | |
| `?` | any symbol | |
| `?*` | the universal ("sigma-star") language (contains all possible strings of any length including the empty string) | |
| `.#.` | string beginning or end (sec. 2.1) | |

Note that expressions that contain the crossproduct (`.x.`) or the composition (`.o.`) operator, describe regular relations rather than regular languages. A regular relation is a mapping from one regular language to another one. Regular languages correspond to simple finite-state automata; regular relations are modelled by finite-state transducers.

In the relation `A .x. B`, we call the first member, `A`, the **upper** language and the second member, `B`, the **lower** language. This choice of words is motivated by the linguistic tradition of writing the result of a rule application underneath the original form. In a cascade of compositions, `R1 .o. R2 ... .o. Rn`, which models a linguistic derivation by rewrite-rules, the upper side of the first relation, `R1`, contains the "underlying lexical form", while the lower side of the last relation, `Rn`, contains the resulting "surface form".

We recognize two kinds of symbols: simple symbols (`a`, `b`, `c`, etc.) and fst pairs (`a:b`, `y:z`, etc.). An fst pair `a:b` can be thought of as the crossproduct of `a` and `b`, the minimal relation consisting of `a` (the upper symbol) and `b` (the lower symbol).

## 2 Parallel Replacement

Conditional parallel replacement denotes a relation which maps a set of $n$ expressions $U_i$ ($i \in [1, n]$) in the upper language into a set of corresponding $n$ expressions $L_i$ in the lower language if, and only if, they occur between a left and a right context ($l_i$, $r_i$).

$$\{ \ U_1 \ \text{->} \ L_1 \ || \ l_1 \_ r_1 \ \} \ , \ \dots \quad [3]$$
$$\dots \ , \ \{ \ U_n \ \text{->} \ L_n \ || \ l_n \_ r_n \ \}$$

Unconditional parallel replacement denotes a similar relation where the replacement is not constraint by contexts.

Conditional parallel replacement corresponds to what Kaplan and Kay (1994) call "batch rules" where a set of rules (replacements) is collected together in a batch and performed in parallel, at the same time, in a way that all of them work on the same input, i.e. not one applies to the output of another replacement.

### 2.1 Examples

Regular expressions based on [3] can be abbreviated if some of the UPPER-LOWER pairs, and/or some of the LEFT-RIGHT pairs, are equivalent. The complex expression:

$$\{ \ \text{a -> b , b -> c || x \_ y} \ \} \ ; \quad [4]$$

which contains multiple replacement in one left and right context, can be written in a more elementary way as two parallel replacements:

$$\{ \ \text{a -> b || x \_ y} \ \}, \{ \ \text{b -> c || x \_ y} \ \}; \ [5]$$

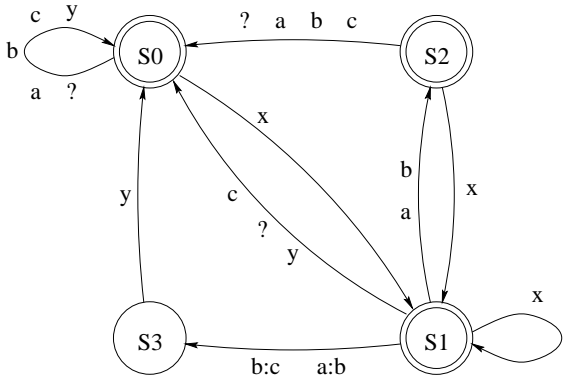

Figure 1: Transducer encoding [4] and [5] *(Every arc with more than one label actually stands for a set of arcs with one label each.)*

Figure 1 shows the state diagram of a transducer resulting from [4] or [5]. The transducer maps the string `xaxayby` to `xaxbyby` following the path `0-1-2-1-3-0-0-0` and the string `xbybyxa` to `xcybyxa` following the path `0-1-3-0-0-0-1-2`.

The complex expression

$$\{ \ \text{a -> b , b -> c || x \_ y , v \_ w} \ \} \ , \quad [6]$$
$$\{ \ \text{a -> c || p \_ q} \ \} \ ;$$

contains five single parallel replacements:

$$\{ \ \text{a -> b || x \_ y} \ \} \ , \quad [7]$$
$$\{ \ \text{a -> b || v \_ w} \ \} \ ,$$
$$\{ \ \text{b -> c || x \_ y} \ \} \ ,$$
$$\{ \ \text{b -> c || v \_ w} \ \} \ ,$$
$$\{ \ \text{a -> c || p \_ q} \ \} \ ;$$

Contexts can be unspecified as in

$$\{ \ \text{a -> b || x \_ y , v \_ \_ w} \ \} \ ; \quad [8]$$

where `a` is replaced by `b` only when occuring between `x` and `y`, or after `v`, or before `w`.

An unspecified context is equivalent to `?*`, the universal (sigma-star) language. Similarly, a specified context, such as `x _ y`, is actually interpreted as `?* x _ y ?*`, that is, implicitly extending the context to infinity on both sides of the replacement. This is a useful convention, but we also need to be able to refer explicitly to the beginning or the end of a string. For this purpose, we introduce a special symbol, `.#.` (Kaplan and Kay, 1994, p. 349).

In the example

$$\{ \ \text{a -> b || .\#.\_ , v \_? ? .\#.} \ \} \ ; \quad [9]$$

`a` is replaced by `b` only when it is at the beginning of a string or between `v` and the two final symbols of a string[1].

### 2.2 Replacement of the Empty String

The language described by the UPPER part of a replacement expression[2]

$$\text{UPPER -> LOWER || LEFT \_ RIGHT} \quad [10]$$

can contain the empty string $\epsilon$. In this case, every string that is in the upper-side language of the relation, is mapped to an infinite set of strings in the lower-side language as the upper-side string can be considered as a concatenation of empty and non-empty substrings, with $\epsilon$ at any position and in any number. E.g.

$$\text{a* -> x || \_} \ ; \quad [11]$$

maps the string `bb` to the infinite set of strings `bb`, `xbb`, `xbxb`, `xbxbx`, `xxbb`, etc., since the language described by `a*` contains $\epsilon$, and the string `bb` can be considered as a result of any one of the concatenations `b⌢b`, $\epsilon$`⌢b⌢b`, `b⌢b⌢`$\epsilon$, `b⌢b⌢`$\epsilon$`⌢`$\epsilon$`⌢b`, $\epsilon$`⌢`$\epsilon$`⌢b⌢b`, etc.

For many practical purposes it is convenient to construct a version of empty-string replacement that allows only one application between any two adjacent symbols (Karttunen, 1995). In order not to confuse the notation by a non-standard interpretation of the notion of empty string, we introduce a special pair of brackets, `[. .]`, placed around the upper side of a replacement expression that presupposes a strict alternation of empty substrings and non-empty substrings of exactly one symbol:

$$\epsilon \frown \text{x} \frown \epsilon \frown \text{y} \frown \epsilon \frown \text{z} \frown \epsilon \frown \ \ldots \qquad [12]$$

---

[1]Note that `.#.` denotes the beginning or the end of a string depending on whether it occurs in the left or the right context.

[2]We describe this topic only for uni-directional replacement from the upper to the lower side of a regular relation, but analogous statements can be made for all other types of replacement mentioned in section 3.

In applying this to the above example, we obtain

```
[. a* .]  -> x || _ ;                    [13]
```

that maps the string **bb** **only** to **xbxbx** since **bb** is here considered exclusively as a result of the concatenation $\epsilon \frown \text{b} \frown \epsilon \frown \text{b} \frown \epsilon$.

If contexts are specified (in opposition to the above example) then they are taken into account.

## 2.3 The Algorithm

### 2.3.1 Auxiliary Brackets

The replacement of one substring by another one inside a context, requires the introduction of auxiliary symbols (e.g. brackets). Kaplan and Kay (1994) motivate this step.

If we would use an expression like

$$l_i \ [U_i \ \text{.x.} \ L_i] \ r_i \qquad [14]$$

to map a particular $U_i$ ($i \in [1, n]$) to $L_i$ when occuring between a left and a right context, $l_i$ and $r_i$, then every $l_i$ and $r_i$ would map substring adjacent to $U_i$.

However, this approach is impossible for the following reason (Kaplan and Kay, 1994): In an example like

```
{ a -> b || x _ x } ;                    [15]
```

where we expect **xaxax** to be replaced by **xbxbx**, the middle **x** serves as a context for both **a**'s. A relation described by [14] could not accomplish this. The middle **x** would be mapped either by an $r_i$ or by an $l_i$ but not by both at the same time. That is why only one **a** could be replaced and we would get two alternative lower strings, **xbxax** and **xaxbx**.

Therefore, we have to use the contexts, $l_i$ and $r_i$, without mapping them. For this purpose we introduce auxiliary brackets $<_i$ after every left context $l_i$ and $>_i$ before every right context $r_i$. The replacement maps those brackets without looking at the actual contexts.

We need separate brackets for empty and non-empty UPPER. If we used the same bracket for both this would mean an overlap of the substrings to replace in an example like $\text{x}>_1<_1\text{a}>_1$. Here we might have to replace $>_1<_1$ after every $<_1\text{a}>_1$ where $<_1$ is part of both substrings. Because of this overlap, we could not replace both substrings in parallel, i.e. at the same time. To make the two replacements sequentially is also impossible in either order, for reasons in detail explained in (Kempe and Karttunen, 1995).

A regular relation describing replacement in context (and a transducer that represents it), is defined by the composition of a set of "simpler" auxiliary relations. Context brackets occur only in intermediate relations and are not present in the final result.

## 2.3.2 Preparatory Steps

Before the replacement we make the following three transformations:

**(1)** Complex regular expressions like [4] are transformed into elementary ones like [5], where every single replacement consists of only one UPPER, one LOWER, one LEFT and one RIGHT expression. E.g.

```
{ [.(a).]  -> b || x _ y } ,
{ [ ] -> c , e -> f || v _ w } ;         [16]
```

would be expanded to

```
{ [.(a).]  -> b || x _ y } ,
{ [ ] -> c || v _ w } ,                   [17]
{ e -> f || v _ w } ;
```

**(2)** Since we have to use different types of brackets for the replacement of empty and non-empty UPPER (cf. 2.3.1), we split the set of parallel replacements into two groups, one containing only replacements with empty UPPER and the other one only with non-empty UPPER. If an UPPER contains the empty string but is not identical with it, the replacement will be added to both groups but with a different UPPER. E.g. [17] would be split into

```
{ a -> b || x _ y } ,
{ e -> f || v _ w } ;                     [18]
```

the group of non-empty UPPER and

```
{ [. .]  -> b || x _ y } ,
{ [ ] -> b || v _ w } ;                   [19]
```

the group of empty UPPER.

**(3)** All empty UPPER of type `[ ]` are transformed into type `[. .]` and the corresponding LOWER are replaced by their Kleene star function. E.g. [19] would be transformed into

```
{ [. .]  -> b || x _ y } ,
{ [. .]  -> c* || v _ w } ;               [20]
```

The following algorithm of conditional parallel replacement will consider all empty UPPER as being of type `[. .]`, i.e. as not being adjacent to another empty string.

### 2.3.3 The Replacement itself

Apart from the previously explained symbols, we will make use of the following symbols in the next regular expressions:

[21]

| | |
|---|---|
| $<_{allE}$ | [ $<_{1E}$ \|...\| $<_{mE}$ ], union of all left brackets for empty UPPER. |
| $>_{allE}$ | [ $>_{1E}$ \|...\| $>_{mE}$ ], union of all right brackets for empty UPPER. |
| $><_{allE}$ | [ $<_{allE}$ \| $>_{allE}$ ] |
| $<_{allNE}$ | [ $<_1$ \|...\| $<_n$ ], union of all left brackets for non-empty UPPER. |
| $>_{allNE}$ | [ $>_1$ \|...\| $>_n$ ], union of all right brackets for non-empty UPPER. |
| $><_{allNE}$ | [ $<_{allNE}$ \| $>_{allNE}$ ] |
| $<_{all}$ | [ $<_{allE}$ \| $<_{allNE}$ ] |
| $>_{all}$ | [ $>_{allE}$ \| $>_{allNE}$ ] |
| $><_{all}$ | [ $<_{all}$ \| $>_{all}$ ] |
| $\diagup$ | Ignore-inside operator. Example: abc$\diagup$x = [abc/x] - [x ?*] - [?* x], inside the string abc, i.e. between a and b and between b and c, all x will be ignored any number of times. |

We compose the conditional parallel replacement of the six auxiliary relations described by Kaplan and Kay (1994) and Karttunen (1995) which are:

```
(1)   InsertBrackets
(2)   ConstrainBrackets
(3)   LeftContext
(4)   RightContext
(5)   Replace
(6)   RemoveBrackets
```
[22]

The composition of these relations in the above order, defines the upward-oriented replacement. The resulting transducer maps UPPER inside an input string to LOWER, when UPPER is between LEFT and RIGHT in the input context, leaving everything else unchanged. Other variants of the replacement operator will be defined later.

For every single replacement $\{\ U_i \rightarrow L_i\ ||\ l_i$ _ $r_i\ \}$ we introduce a separate pair of brackets $<_i$ and $>_i$ with $i \in [1E...mE]$ if UPPER is identical with the empty string and $i \in [1...n]$ if UPPER does not contain the empty string. A left bracket $<_i$ indicates the end of a complete left context. A right bracket $>_i$ marks the beginning of a complete right context.

We define the component relations in the following way. Note that UPPER, LOWER, LEFT and RIGHT ($U_i$, $L_i$, $l_i$ and $r_i$) stand for regular expressions of any complexity but restricted to denote regular languages. Consequently, they are represented by networks that contain no fst pairs.

### (1) InsertBrackets

```
[ ]   <-   ><_all
```
[23]

The relation inserts instances of all brackets on the lower side (everywhere and in any number and order).

### (2) ConstrainBrackets

```
  ˜$[ >_allE   [ >_allNE ] ]
& ˜$[ <_allE   [ >_all ] ]
& ˜$[ <_allNE  [ <_allE | >_all ] ]
```
[24]

The language does not apply to single brackets but to their types and allows them to be only in the following order:

```
>_allNE*  >_allE*  <_allE*  <_allNE*
```
[25]

The composition of the steps (1) and (2) invokes this constraint, which is necessary for the following reasons:

If we allowed sequences like $<_3 U_3 <_1 >_3 U_1 >_1$ we would have an overlap of the two substrings $<_3 U_3 >_3$ and $<_1 U_1 >_1$ which have to be replaced. Here, either $U_1$ or $U_3$ could be replaced but not both at the same time.

If we permitted sequences like $>_{1E} <_2 <_{1E} U_2 >_2$ we would also have an overlap of the two replacements which means we could either replace $<_2 U_2 >_2$ or $>_{1E} <_{1E}$ but not both.

### (3) LeftContext

```
λ_1 & ... & λ_n
```
[26]

```
for all i ∈ [1E...mE, 1...n]  ,  λ_i =
    ˜$[ ˜[l_i ↙>_{all}]   (><_all − <_i)*   <_i ]
  & ˜$[ [l_i ↙>_{all}]    (><_all − <_i)*   ˜<_i ]
```

The constraint forces every instance of a left bracket $<_i$ to be immediately preceded by the corresponding left context $l_i$ and every instance of $l_i$ to be immediately followed by $<_i$, ignoring all brackets that are different from $<_i$ inbetween, and all brackets ($<_i$ included) inside $l_i$ (↙). We separately make the constraints $λ_i$ for every $<_i$ and $l_i$ and then intersect them in order to get the constraint for all left brackets and contexts.

### (4) RightContext

```
ρ_1 & ... & ρ_n
```
[27]

```
for all i ∈ [1E...mE, 1...n]  ,  ρ_i =
    ˜$[ >_i   (><_all − >_i)*   ˜[r_i ↙><_all] ]
  & ˜$[ ˜>_i  (><_all − >_i)*   [r_i ↙><_all] ]
```

The constraint relates instances of right brackets $>_i$ and of right contexts $r_i$, and is the mirror image of step (3). We derive it from the left context constraint by reversing every right context $r_i$, before making the single constraints $λ_i$ (not $ρ_i$) and reversing again the result after having intersected all $λ_i$.

### (5) Replace

```
[ N R ]* N
```
[28]

The relation maps every bracketed UPPER, $<_i U_i >_i$ for non-empty UPPER and $>_i <_i$ for empty UPPER, to the corresponding bracketed LOWER, $<_i L_i >_i$, leaving everything else unchanged.

The term $N$ in [28] means a string that does not contain any bracketed UPPER:

$$N = N_{1E}\ \&...\&\ N_{mE}\ \&\ N_1\ \&...\&\ N_n$$
[29]

A particular bracketed empty UPPER $>_i <_i$ is excluded from the corresponding $N_i$ ($i \in [1E, mE]$) by

$$N_i = ˜\$[>_i\ [><_{allE} − >_i − <_i]*\ <_i]$$
[30]

and a bracketed non-empty UPPER $<_i U_i >_i$ is excluded from the corresponding $N_i$ ($i \in [1, n]$) by

$$N_i = ˜\$[<_i\ [<_{allNE} − <_i]*$$
$$U_1 ↙>_{all}\ [>_{allNE} − >_i]*\ >_i]$$
[31]

The term $R$ in expression [28] abbreviates a relation that maps any bracketed UPPER to the corresponding bracketed LOWER. It is the union of all single $R_i$ relations mapping all occurrences of one $U_i$ (empty and non-empty) to the corresponding $L_i$:

$$R = R_{1E}\ |...|\ R_{mE}\ |\ R_1\ |...|\ R_n$$
[32]

**The replacement $R_i$ of non-empty UPPER** $U_i$ ($i \in [1, n]$) is performed by:

$$<_i\ \ [\ [U_i ↙>_{all}].x.[L_i ↙>_{all}]\ ]\ \ >_i$$
[33]

To illustrate this: Suppose we have a set of replacements containing among others

```
a -> b || x _ y ;
```
[34]

This particular replacement is done by mapping inside an input string every substring that looks like (underlined part)

$$...x >_2 >_{1} >_{1E} <_{1E} <_2\ \underline{<_1 a >_1}\ >_2 >_{1E} <_{1E} <_1 <_2 y...$$
[35]

using the brackets $<_1$ and $>_1$ to a substring (underlined part)

**The replacement $\mathcal{R}_i$ of empty UPPER $U_i$** ($i \in [1E, mE]$) is performed by:

```
[ 0.x.[[><allE - <i] | [<allINE]] ]* [37]
[>i.x.<i] [ 0.x.[Li _/>><all]] [<i.x.>i]
[ 0.x.[[><allE - >i] | [>allINE]] ]*
```

In the following example we replace the empty $U_{2E}$ by $L_{2E}$. Suppose we have in total one replacement of non-empty UPPER and two of empty UPPER, one of which is

```
[. .]  -> b || x _ y ;        [38]
```

This replacement is done by mapping inside a string every substring that looks like (underlined part)

...x $>_1>_{1E}$  <u>$>_{2E}$ $<_{2E}$</u>  $<_{1E}<_1$ y... [39]

using the brackets $>_{2E}<_{2E}$ into a substring (underlined part)

...x $>_1>_{1E}$ <u>$[>_{2E} | >_{1E} | <_{1E} | <_1]$*</u> [40]
<u>$<_{2E}$b$>_{2E}$</u>
<u>$[>_1 | >_{1E} | <_{1E} | <_{2E}]$*</u> $<_{1E}<_1$ y...

The occurrence of exactly one bracket pair $>_{iE}$ and $<_{iE}$ between a left and a right context, actually corresponds to the definition of a (single) empty string expressed by [. .] (cf. sec. 2.2).

The brackets $[>_{2E} | >_{1E} | <_{1E} | <_1]$ and $[>_1 | >_{1E} | <_{1E} | <_{2E}]$ in [40] are inserted on the lower side any number of times (including zero), i.e. they exist optionally, which makes them present if checking for the left or right context requires them, and absent if they are not allowed in this place. This set of brackets does not contain those ones used for the replacement, $>_i<_i$, because if we later check for them we do not want this check to be always satisfied but only when the specified contexts are present, in order to be able to confirm or to cancel the replacement *a posteriori*.

This set of optionally inserted brackets equally does not contain those which potentially could be used for the replacement of adjacent non-empty strings, i.e. $>_{allNE}$ on the left and $<_{allNE}$ on the right side of the expression. Otherwise, checking later for the legitimacy of the adjacent replacements would no longer be possible.

**(6) RemoveBrackets**

```
><all  ->  [ ]               [41]
```

The relation eliminates from the lower-side language all brackets that appear on the upper side.

## 3   Variants of Replacement

### 3.1   Application of context constraints

We distinguish four ways how context can constrain the replacement. The difference between them is where the left and the right contexts are expected, on the upper or on the lower side of the relation, i.e. LEFT and RIGHT contexts can be checked before or after the replacement.

We obtain these four different applications of context constraints (denoted by ||, //, \\ and

\/) by varying the order of the auxiliary relations (steps (3) to (5)) described in section *2.3.3* (cf. [22]) :

(a) Upward-oriented

```
{ U1 -> L1 || l1 _ r1 } , ...
... , { Un -> Ln || ln _ rn }   [42]
...LeftContext .o. RightContext .o. Replace...
```

(b) Right-oriented

```
{ U1 -> L1 // l1 _ r1 } , ...        [43]
...RightContext .o. Replace .o. LeftContext...
```

(c) Left-oriented

```
{ U1 -> L1 \\ l1 _ r1 } , ...        [44]
...LeftContext .o. Replace .o. RightContext...
```

(d) Downward-oriented

```
{ U1 -> L1 \/ l1 _ r1 } , ...        [45]
...Replace .o. LeftContext .o. RightContext...
```

The versions (a) to (c) roughly correspond to the three alternative interpretations of phonological rewrite rules discussed in Kaplan and Kay (1994). The upward-oriented version corresponds to the simultaneous rule application; the right- and left-oriented versions can model rightward or leftward iterating processes, such as vowel harmony and assimilation.

In the downward-oriented replacement the operation is constrained by the lower (left and right) context. Here the $U_i$ get mapped to the corresponding $L_i$ just in case they end up between $l_i$ and $r_i$ in the output string.

### 3.2   Inverse, bidirectional and optional replacement

Replacement as described above, ->, maps every $U_i$ on the upper side unambiguously to the corresponding $L_i$ on the lower side but not vice versa. A $L_i$ on the lower side gets mapped to $L_i$ or $U_i$ on the upper side.

The inverse replacement, <-, maps unambiguously from the lower to the upper side only. The bidirectional replacement, <->, is unambiguous in both directions.

Replacements of all of these three types (directions) can be optional, (->) (<-) (<->), i.e. they are either made or not. We define such a relation by changing $\mathcal{N}$ (the part not containing any bracketed UPPER) in expression [28] into ?* that accepts every substring:

```
[ ?* R]* ?*                  [46]
```

Here an $U_i$ is either mapped by the corresponding $\mathcal{R}_i$ contained in $\mathcal{R}$ (cf. [32]) and therefore replaced by $L_i$, or it is mapped by ?* and not replaced.

## 4   A Practical Application

In this section we illustrate the usefulness of the replace operator using a practical example.

We show how a lexicon of French verbs ending in *-ir*, inflected in the present tense subjunctive mood, can be derived from a lexicon containing the corresponding present indicative forms. We assume here that irregular verbs are encoded separately.

It is often proposed that the present subjunctive of *-ir* verbs be derived, for the most basic case, from a stem in *-iss-* (e.g.: fin*ir*/fin*iss*) rather than from a more general root (e.g.: fin(i)) because once this stem is assumed, the subjunctive ending itself becomes completely regular:

| (that I finish) | (that I run) |
|---|---|
| *que je finiss-e* | *que je cour-e* |
| *que tu finiss-es* | *que tu cour-es* |
| ..... | ..... |
| *que ils finiss-ent* | *que ils cour-ent* |

The algorithm we propose here, is straightforward: We first derive the present subjunctive stem from the third person plural present indicative (e.g.: fin*iss*, cour), then append the suffix corresponding to the given person and number.

The first step can be described as follows:

```
define LETTER :                              [47]
    a | b | c | d | .... ;

define TAG :                                 [48]
    SubjP|...|SG|...|P3|...|Verb|... ;

define StemRegular :                         [49]
    [ [. .] <-> IndP PL P3 Verb || LETTER _ TAG ]
        .o.
    [ LexInd  TAG+ ]
        .o.
    [ e n t <-> SUFF || _ TAG ] ;
```

The first transducer in [49] inserts the tags of the third person plural present indicative between the word and the tags of the actually required subjunctive form. The second transducer in [49] which is an indicative lexicon of *-ir* verbs, concatenated with a sequence of at least one tag, provides the indicative form and keeps the initial subjunctive tags. The last transducer in [49] replaces the suffix *-ent* by the symbol SUFF. E.g.:

```
finir__________________SubjP_PL_P2_Verb
finir__IndP_PL_P3_Verb__SubjP_PL_P2_Verb
finissent______________SubjP_PL_P2_Verb
finiss_SUFF____________SubjP_PL_P2_Verb
```

To append the appropriate suffix to the subjunctive stem, we use the following transducer which maps the symbol SUFF to a suffix and deletes all tags:

```
define Suffix :                              [50]
    [ { SUFF -> e      || _ TAG* SG [P1|P3] },
      { SUFF -> e s    || _ TAG* SG P2 },
      { SUFF -> i o n s || _ TAG* PL P1 },
      { SUFF -> i e z   || _ TAG* PL P2 },
      { SUFF -> e n t   || _ TAG* PL P3 } ]
        .o.
    [ TAG -> [ ] ] ;
```

The complete generation of subjunctive forms can be described by the composition:

```
define LexSubjP :                            [51]
    StemRegular  .o.  Suffix ;
```

The resulting (single) transducer LexSubjP represents a lexicon of present subjunctive forms of French verbs ending in *-ir*. It maps the infinitive of those verbs followed by a sequence of subjunctive tags, to the corresponding inflected surface form and vice versa.

All intermediate transducers mentioned in this section will contribute to this final transducer but will themselves disappear.

The regular expressions in this section could also be written in the two-level formalism (Koskenniemi, 1983). However, some of them can be expressed more conveniently in the above way, especially when the replace operator is used.

E.g., the first line of [49], written above as:

```
[. .] <-> IndP PL P3 Verb || LETTER _ TAG     [52]
```

would have to be expressed in the two-level formalism by four rules:

```
0:IndP <=> LETTER    _ (:PL)(:P3)(:Verb) TAG; [53]
0:PL   <=> LETTER (:IndP) _ (:P3)(:Verb) TAG;
0:P3   <=> LETTER (:IndP)(:PL) _ (:Verb) TAG;
0:Verb <=> LETTER (:IndP)(:PL)(:P3) _    TAG;
```

Here, the difficulty comes not only from the large number of rules we would have to write in the above example, but also from the fact that writing one of these rules requires to have in mind all the others, to avoid inconsistencies between them.

## Acknowledgements

This work builds on the research by Ronald Kaplan and Martin Kay on the finite-state calculus and the implementation of phonological rewrite rules (1994).

Many thanks to our collegues at PARC and RXRC Grenoble who helped us in whatever respect, particularly to Annie Zaenen, Jean-Pierre Chanod, Marc Dymetman, Kenneth Beesley and Anne Schiller for helpful discussion on different topics, and to Irene Maxwell for correcting the paper.